\documentclass[10pt,twocolumn,letterpaper]{article}

\usepackage[pagenumbers]{cvpr} % To force page numbers, e.g. for an arXiv version

\usepackage{graphicx}
\usepackage{amsmath}
\usepackage{amssymb}
\usepackage{booktabs}

\usepackage{cuted}
\usepackage{multirow}

\usepackage[pagebackref,breaklinks,colorlinks]{hyperref}
\usepackage{mathtools}

\usepackage[capitalize]{cleveref}
\crefname{section}{Sec.}{Secs.}
\Crefname{section}{Section}{Sections}
\Crefname{table}{Table}{Tables}
\crefname{table}{Tab.}{Tabs.}

\newcommand{\norm}[1]{\left\lVert#1\right\rVert}

\def\cvprPaperID{3247} % *** Enter the CVPR Paper ID here
\def\confName{CVPR}
\def\confYear{2022}

\begin{document}

%%%%%%%%% TITLE - PLEASE UPDATE
\title{Interpretable part-whole hierarchies and conceptual-semantic relationships in neural networks}

% agGLOMmerate: interpretable part-whole hierarchies and conceptual-semantic relationships in neural networks
% HeraNeT: HiERAchical neural Network

\author{Nicola Garau, Niccol\'o Bisagno, Zeno Sambugaro, and Nicola Conci\\
University of Trento - Department of Information Engineering and Computer Science - DISI\\
Via Sommarive, 9, 38123 Povo, Trento TN\\
{\tt\small nicola.garau,niccolo.bisagno,zeno.sambugaro,nicola.conci@unitn.it}
}

\maketitle

\vspace*{-100pt}
\begin{strip}
    \centering
    \includegraphics[width=0.95\textwidth]{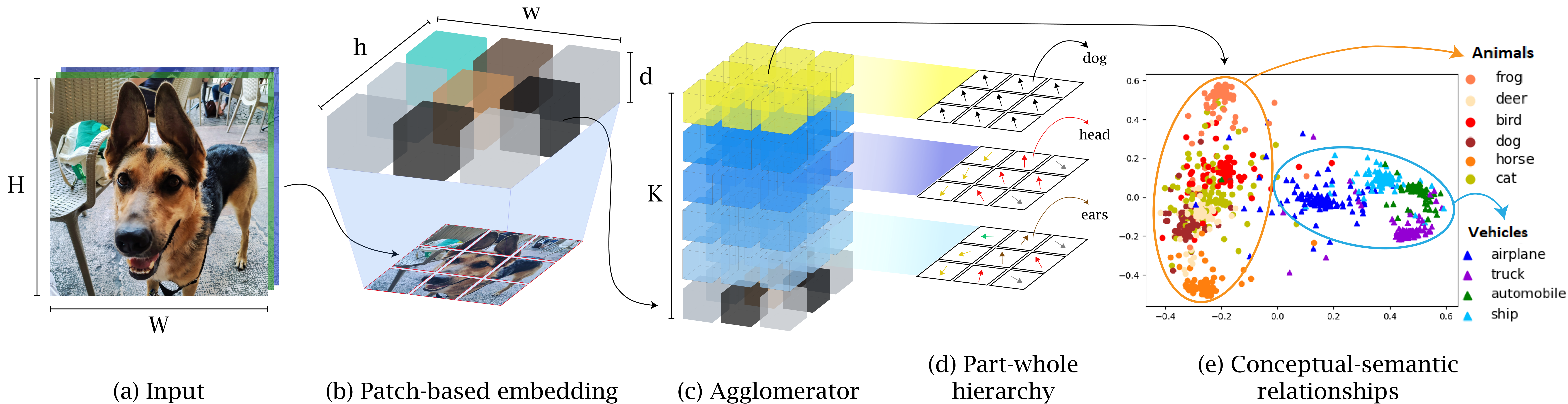}
    \captionof{figure}{ \textbf{[Better seen in color]}. Overview of the proposed solution. Our Agglomerator is a novel architecture for vision applications, in which column structure (c) mimics hyper-columns typical of the human visual cortex \cite{hawkins2021thousand}. The input data (a) is fed to the columns using a patch-based embedding (b). The Agglomerator architecture iteratively routes the information across its structure, creating a neural representation of each image, similar to neural fields \cite{mildenhall2020nerf}. In the neural representation, \textit{part-whole} hierarchies (d) emerge at different levels of the columns. The same column can represent the same patch of the image with different levels of abstraction (e.g., the ears, the head, and the dog) corresponding to each level in the column. Neighbor columns agree on a \textit{part} representation (e.g ears, head) at lower levels, ideally representing the same \textit{whole} (e.g. dog) at the top level. The resulting feature space represents the \textit{conceptual-semantic relationships} between data (e) resembling the human hierarchical organization \cite{miller1995wordnet}. Samples belonging to the same super-class (e.g., animals, vehicles) are clustered together, with conceptually close categories (e.g., birds and airplanes) represented on the edge of the super-classes.}
    \label{fig:teaser}
\end{strip}
\vspace*{-20pt} 

%%salva versione prima di camera ready

%%%%%%%%% ABSTRACT
\begin{abstract}
\vspace*{-12pt} 
   Deep neural networks achieve outstanding results in a large variety of tasks, often outperforming human experts. However, a known limitation of current neural architectures is the poor accessibility to understand and interpret the network response to a given input. This is directly related to the huge number of variables and the associated non-linearities of neural models, which are often used as black boxes. When it comes to critical applications as autonomous driving, security and safety, medicine and health, the lack of interpretability of the network behavior tends to induce skepticism and limited trustworthiness, despite the accurate performance of such systems in the given task. Furthermore, a single metric, such as the classification accuracy, provides a non-exhaustive evaluation of most real-world scenarios.
   In this paper, we want to make a step forward towards interpretability in neural networks, providing new tools to interpret their behavior.
   We present Agglomerator, a framework capable of providing a representation of part-whole hierarchies from visual cues and organizing the input distribution matching the conceptual-semantic hierarchical structure between classes. We evaluate our method on common datasets, such as SmallNORB, MNIST, FashionMNIST, CIFAR-10, and CIFAR-100, providing a more interpretable model than other state-of-the-art approaches.
\end{abstract}

%%%%%%%%% BODY TEXT
\vspace*{-30px}
\section{Introduction}
\label{sec:intro}

%parte avanzata riutilizzabile
% Starting from visual cues, humans can identify part-whole relationships between elements in a scene. If deep learning networks can internally identify such relationships, their representation is not interpretable by humans.
%   Moreover, people are naturally able to analyze objects, delivering aggregate information such as their definition and organization in conceptual-semantic categories. Common deep learning networks can not perform such categorization, resulting in samples with close conceptual-semantic and lexical relations being far away from each other in the feature space. 

The extensive adoption of neural networks and, in general, learning models has been raising concerns regarding our chances, as humans, to explain their behavior.
Interpretability would be a highly desirable feature for neural networks, especially in those applications like autonomous driving \cite{grigorescu2020survey}, healthcare \cite{miotto2018deep}, and finance \cite{sezer2020financial}, where safety, life, and security are at stake.

Deep neural networks have achieved superhuman performances in many domains, from computer vision \cite{lecun2015deep, he2016deep} to natural language processing \cite{vaswani2017attention,devlin2018bert}, and data analysis \cite{sezer2020financial}. However, the achieved performances have come at the expense of model complexity, making it difficult to interpret how neural networks work \cite{linardatos2021explainable}. These neural networks are usually deployed as ''black boxes'', with millions of parameters to be tuned, mostly according to experience and rule of thumb. Interpreting how a trainable parameter in the network setup directly affects the desired output from a given input has nearly zero chances.

According to the literature, interpretability is defined as \textit{“the degree to which a human can understand the cause of a decision”}\cite{miller2019explanation}. When a machine learning model reaches high accuracy on a task such as classification and prediction, can we trust the model without understanding why such a decision has been taken? The decision process is complex and we tend to evaluate the performance of a system in solving a given task using metrics computed at the end of the processing chain. While single metrics, such as the classification accuracy, reach super-human results, they provide an incomplete description of the real-world task \cite{doshi2017towards}.
%For example, when dealing with an image classification problem, the learning model can tell the class the represented object belongs to. Thus, we know \textbf{what} has been predicted by the network, but we have little understanding about \textbf{why} we have obtained such prediction \cite{molnar2019}. 
As humans, when looking at an object that has eyes and limbs, we can infer via reasoning and intuition that these are elements (parts) that belong to the same entity (whole) \cite{biederman1987recognition}, say an animal, and we can explain and motivate why such decision is taken, generally based on past experiences, beliefs and attitude \cite{albarracin2000cognitive}. Moreover, even in presence of an animal never seen before, we can probably tell from the visual features, our frames of reference \cite{hawkins2021thousand} and our hierarchical organization of objects in the world \cite{miller1995wordnet} whether it is a fish or a mammal. We would like neural networks to display the same behavior, so that objects that are \textit{close} in the conceptual-semantic and lexical relations are adjacent in the feature space as well (as shown in Fig. \ref{fig:teaser}\textcolor{red}{e}). By doing so, it would be intuitive to identify hierarchical relations between samples and how the model has learned to build a topology describing each sample.
Consequently, we can agree on the definition of interpretability in deep learning as the \textit{“extraction of relevant knowledge from a machine-learning model concerning relationships either contained in data or learned by the model”} \cite{murdoch2019definitions}.

In the image classification field, available techniques, such as transformers \cite{vaswani2017attention,dosovitskiy2020image,devlin2018bert}, neural fields \cite{mildenhall2020nerf}, contrastive learning representation \cite{chen2020simple}, distillation \cite{hinton2015distilling} and capsules \cite{sabour2017dynamic}, have achieved state-of-the-art performances, introducing a number of novelties, such as powerful attention-based features and per-patch analysis, positional encoding, similarity-based self-supervised pre-training, model compression and deep modeling of part-whole relationships.
Taken as standalone, these methods have contributed to improving the interpretability of networks, while still lacking direct emphasis on either data relationships \cite{vaswani2017attention,dosovitskiy2020image,devlin2018bert,chen2020simple,mildenhall2020nerf} (e.g. conceptual-semantic relationships) or model-learned relationships \cite{sabour2017dynamic,hinton2015distilling} (e.g. part-whole relationships). Retrieving part whole hierarchy is not a new task per se, as it has been exploited in different research areas as scene parsing \cite{bear2020learning,deng2020generative} and multi-level scene decomposition \cite{zhu2007stochastic,hong2021ptr}. Instead of aiming at learning the part-whole hierarchy as the final goal of our architecture, we focus on learning the part-whole representation as a mean to interpret the network behavior at different levels.\\
In \cite{hinton2021represent}, a concept idea on how to represent part-whole hierarchies in neural networks is introduced, which attempts to merge the advantages of the above state-of-the-art frameworks into a single theoretical system (known as \textit{GLOM}). GLOM aims at mimicking the human ability in learning to parse visual scenes. Inspired by the theoretical concepts described in \cite{hinton2021represent, hawkins2021thousand}, we build a working system, called Agglomerator, which achieves part-whole agreement \cite{hinton1990mapping} at different levels of the model (\textit{relationships learned by the model}) and hierarchical organization of the feature space (\textit{relationships contained in data}), as shown in Fig. \ref{fig:teaser}.

Our contribution is summarised as follows:
\begin{itemize}
    \item we introduce a novel model, called Agglomerator\footnote{The code and the pre-trained models can be found at \href{https://github.com/mmlab-cv/Agglomerator}{https://github.com/mmlab-cv/Agglomerator}}, mimicking the functioning  of the cortical columns in the human brain \cite{hawkins2017theory};
    \item we explain how our architecture provides interpretability of \textit{relationships learned by the model}, specifically part-whole relationships;
    \item we show how our architecture provides interpretability of \textit{relationships contained in data}, namely the hierarchical organization of the feature space;
    %closely resembling human lexical similarities\cite{miller1995wordnet}.
    \item we provide results outperforming or on par with current methods on multiple common datasets, such as SmallNORB \cite{lecun2004learning}, MNIST \cite{lecun1998gradient}, FashionMNIST \cite{xiao2017/online}, CIFAR-10 and CIFAR-100 \cite{krizhevsky2009learning}, also relying on fewer parameters.
    %\item we show that our model relies on fewer parameters and can generalize to multiple datasets.
\end{itemize}

\section{Related work}
\label{sec:relatedwork}

\textit{Convolutional Neural Networks (CNNs)} \cite{he2016deep,simonyan2014very} have risen to a prominent role in computer vision when they started to outperform the existing literature in the image classification task of the ImageNet challenge \cite{krizhevsky2012imagenet}. The convolution operator can effectively describe spatially-correlated data resulting in a feature map, while the pooling operation down-samples the obtained feature map by summarizing the presence of certain features in patches of the image. The pooling operation in CNNs has been the subject of criticism since it does not preserve the information related to the part-whole relationship \cite{sitzmann2019scene} between features belonging to the same object \cite{sabour2017dynamic}. 

\textit{Transformers} \cite{liu2021swin,dosovitskiy2020image,khan2021transformers} have proven able to outperform CNNs, thanks to their ability to encode powerful features using self-attention and patch-based analysis of images. Multi-headed transformers \cite{devlin2018bert} require the query, key, and value weights to be trained differently for each head, which is more costly than training a CNN. The main advantage compared to CNNs is the ability of the multiple heads to combine information from different locations in the image with fewer losses than the pooling operation \cite{lee2019set}. However, when compared with CNNs, Transformer-like models usually require intensive pre-training on large datasets, to achieve state-of-the-art performances.

\textit{Multi Layer Perceptrons (MLPs)} \cite{tolstikhin2021mlp,li2021convmlp} are characterised by fully connected layers, in which each node is connected to every other possible node of the next layer. Even though they are easier to train and have simpler architecture compared to CNNs, the fully connected layers may cause the network to grow too fast in size and number of parameters, not allowing powerful scalability. MLPs have experienced a resurgence, thanks to patch-based approaches \cite{tolstikhin2021mlp,li2021convmlp}, that allowed reaching state-of-the-art performances. They can also be seen as 1x1 convolutions \cite{hinton2021represent,tolstikhin2021mlp,li2021convmlp}, which do not require the pooling operation.

\textit{Capsules networks} \cite{sabour2017dynamic,hinton2018matrix,kosiorek2019stacked,ribeiro2020capsule,mukhometzianov2018capsnet,mazzia2021efficient} try to mimic the way the human brain creates a parse tree of parts and wholes by dynamically allocating groups of neurons (capsules) that can model objects at different levels of the part-whole hierarchy. The routing algorithm determines which capsules are activated to describe an object in the image, with lower-level capsules describing the parts (e.g. eyes and limbs), and higher-level capsules describing wholes (e.g. mammals and fish). While effectively routing information from different locations in the image, activated capsules cannot describe every single possible object in the image, thus limiting their effectiveness on more complex datasets (e.g. ImageNet, CIFAR-100), while achieving state-of-the-art results on simpler ones (e.g. MNIST). While part-whole hierarchies have been investigated in other fields like scene parsing \cite{bear2020learning,deng2020generative} and multi-level scene decomposition \cite{zhu2007stochastic,hong2021ptr}, capsule networks aim at building an internal representation of the hierarchy, which allows for better interpretability of the final task (e.g. classification).

There has been a recent push toward the so-called biologically inspired Artificial Intelligence (AI) \cite{hole2021thousand,hawkins2021thousand}, which tries to build deep learning networks able to mimic the structure and functions of the human brain. In \cite{hawkins2021thousand}, the authors propose a column-like structure, similar to hyper-columns typical of the human neocortex. In \cite{van2021disentangling}, the authors build upon cortical columns implemented as separate neural networks called Cortical Column Networks (CCN). Their framework aims at representing part-whole relationships in scenes to learn object-centric representations for classification.  

The author in \cite{hinton2021represent} proposes a conceptual framework, called GLOM, based on inter-connected columns, each of which is connected to a patch of the image and is composed of auto-encoders stacked in levels. Weights sharing among MLP-based \cite{li2021convmlp} auto-encoders allows for an easily trainable architecture with fewer weights, while knowledge distillation \cite{hinton2015distilling} allows for a reduction of the training parameters. The patch-based approach combined with the spatial distribution of columns allows for a sort of positional encoding and viewpoint estimation similarly to what is used in \textit{neural fields} \cite{mildenhall2020nerf,sitzmann2019scene}.
At training time, the author recommends that GLOM should be trained using a contrastive loss function \cite{chen2020simple}. This procedure, combined with a Transformer-like self-attention \cite{vaswani2017attention} mechanism on each layer of the columns, aims at reaching a consensus between columns. Routing the information with layer-based attention and stacked autoencoders would theoretically allow GLOM to learn a different level of abstraction of the input at a different location and level in the columns, creating a part-whole structure with a richer representation if compared to capsule networks \cite{sabour2017dynamic}.

While GLOM is presented in \cite{hinton2021represent} more as an intuition rather than a proper architecture, in this work we develop its foundational concepts and turn them into a fully working system, with application to image classification.

\section{Method}
\label{sec:method}

\begin{figure}
    \centering
    \includegraphics[width=.45\textwidth]{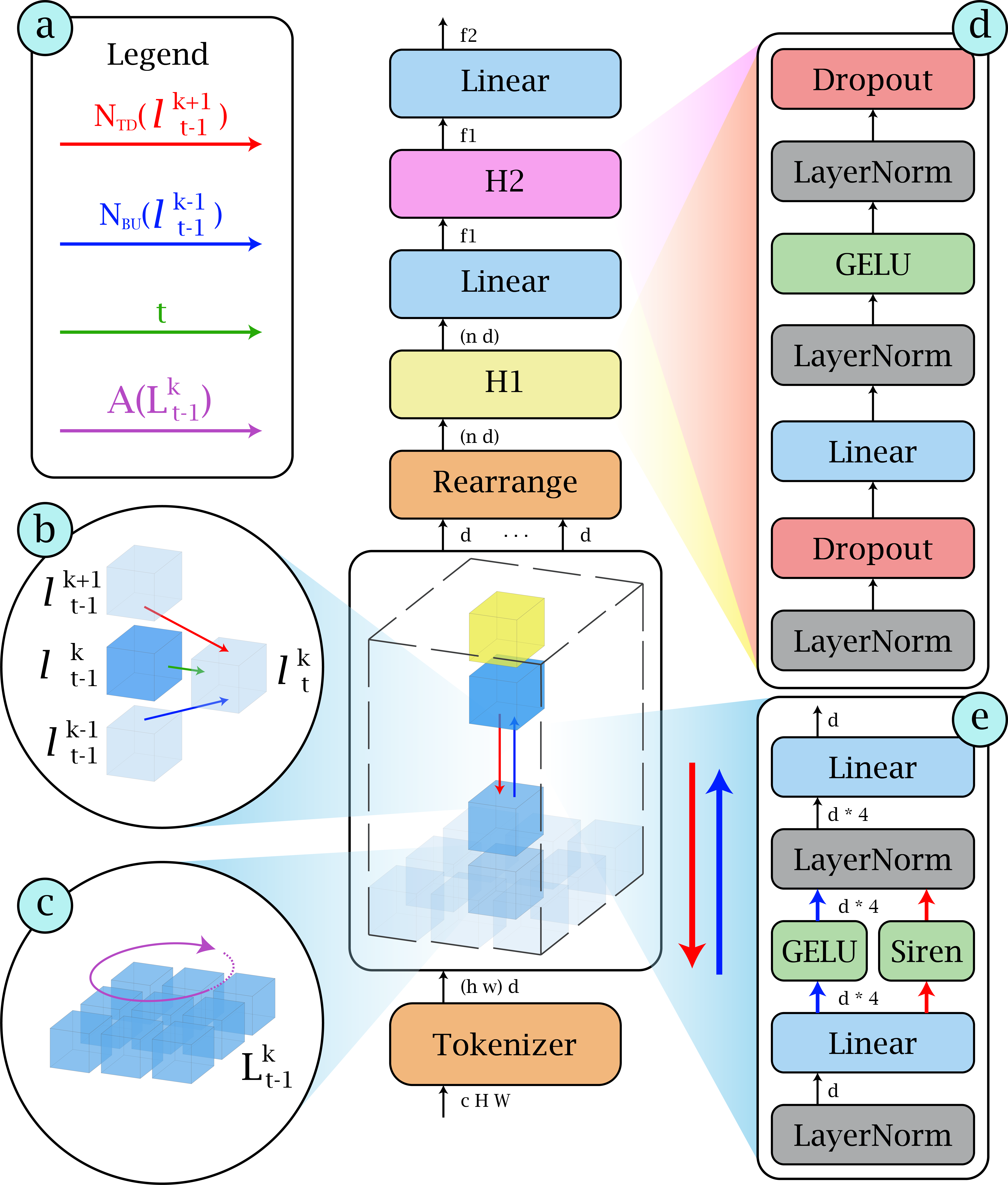}
    \caption{\textbf{[Better seen in color]}. Architecture of our Agglomerator model (center) with information routing (left) and detailed structure of building elements (right). Each \textit{cube} represents a level $l_{t}^{k}$. \textbf{Left:} (a) legend of the arrows in the figure, representing the top-down network $N_{TD}(l_{t-1}^{k+1})$, the bottom-up network $N_{BU}(l_{t-1}^{k-1})$, attention mechanism $A(L_{t-1}^{k})$ and time step $t$. (b) Contribution to the value of level $l_{t}^{k}$ given by $l_{t-1}^{k}$, $N_{TD}(l_{t-1}^{k+1})$ and $N_{BU}(l_{t-1}^{k-1})$. (c) The attention mechanisms $A(L_{t-1}^{k})$ share information between $l_{t-1}^{k} \in L_{t-1}^{k}$. \textbf{Center:} bottom to top, the architecture consists of the Tokenizer module, followed by the columns $C(h,w)$, with each level $l_{t}^{k}$ connected to the neighbors with $N_{TD}(l_{t-1}^{k+1})$ and $N_{BU}(l_{t-1}^{k-1})$. On top of the structure, the contrastive $H1$ and cross entropy $H2$ heads. \textbf{Right:} (d) structure of heads $H1$ and $H2$. (e) Structure of the top-down network $N_{TD}(l_{t-1}^{k+1})$ and the bottom-up network $N_{BU}(l_{t-1}^{k-1})$.}
    \label{fig:architecture}
    \vspace*{-10px}
\end{figure}

% To better explain the network displayed in Fig. \ref{fig:architecture}, we define:
% \begin{itemize}
%     \item a column $C(h,w)$ which consists of a stack of auto-encoders at location $(h,w)$at location $(h,w)$at location $(h,w)$.
%     \item a level $l$ of column $C(h,w)$. Each level $l_{t}^{k}$ of the column is an embedding vector representation of size $d$ and it corresponds to the part-whole hierarchy at a certain level of abstraction $k$ at time $t$. $l_{t}^{k-1}$ and $l_{t}^{k}$ represent consecutive levels, where $l_{t}^{k-1}$ represents a \textit{part} of the \textit{whole} $l_{t}^{k}$. All the levels $l_{t}^{k}$ belong to the same layer $L_{t}^{k}$. %The same relationship withholds with levels $l$ and $l+1$.
%     \item each auto-encoder connects a level $l_{t}^{k}$ to an adjacent level $l_{t}^{k-1}$ or $l_{t}^{k+1}$ using a $N_{TD}(l_{t}^{k})$ top-down decoder and $N_{BU}(l_{t}^{k})$ bottom-up encoder.
%     \item attention $A(L_{t}^{k})$
% \end{itemize}

The framework we propose aims at replicating the column-like pattern, similar to hyper-columns typical of the human visual cortex \cite{hawkins2021thousand}. An overview is shown in Fig. \ref{fig:teaser}.

Agglomerator brings together concepts and building blocks from multiple methods, such as CNNs \cite{li2021convmlp}, transformers \cite{vaswani2017attention,dosovitskiy2020image,devlin2018bert}, neural fields \cite{mildenhall2020nerf}, contrastive learning representation \cite{chen2020simple}, distillation \cite{hinton2015distilling}, and capsules \cite{sabour2017dynamic}.
Here, we introduce the mathematical notation needed to explain the details of the main building blocks of the architecture.

Each input image is transformed into a feature map divided into $N=h\times w$ patches.
The $n$-th patch, with $n\in \{1,\dots, N\}$ is fed to the corresponding column $C_n(h,w)$, spatially located at coordinates $(h,w)$. The subscript $n$ is omitted in the next equations for better readability.
As shown in Fig. \ref{fig:architecture}, each column $C(h,w)$ consists of $K$ embedding levels $\{l_t^{(h,w),k} \mid k=0,\dots,K\}$ connected by a stack of auto-encoders at location $(h,w)$ at time $t\in\{0,\dots,t-1,t,t+1,\dots,T\}$, as suggested in \cite{hinton2021represent}. The superscript $(h,w)$ is omitted in the next instances of $l_{t}^{k}$ for better readability.
Each level $l_{t}^{k}$ of the column is an embedding vector representation of size $d$. Levels $l_{t}^{k-1}$ and $l_{t}^{k}$ represent consecutive levels; $l_{t}^{k-1}$ represents a \textit{part} of the \textit{whole} $l_{t}^{k}$. We indicate as $l_{t}^{k} \in L_{t}^{k}$ all the levels $l_{t}^{k}$ in all columns $C(h,w)$ sharing the same $k$ value and belonging to the same layer $L_{t}^{k}$. Being $K$ the last layer of our architecture at the last time step $T$, it is represented as $L_{T}^{K}$.

\subsection{Patches embedding}
\label{sec:embedding}

At the embedding stage, as in \cite{li2021convmlp}, we apply a convolutional Tokenizer to extract the feature map of each image of size $H\times W$ pixels, which provides a richer representation compared to the original image. Following the implementation in \cite{li2021convmlp}, the obtained feature map has size $h \times w \times d$ where $h = H/4$ and $w=W/4$. We then embed each of the $n$ $d$-dimensional embedding vectors into the bottom levels $l_{t}^{0} \in L_{t}^{0}$ at the corresponding coordinates $(h,w)$ of the corresponding column $C(h,w)$. Feeding the $n$-th each patch to a spatially located column $C(h,w)$ resembles the positional encoding of neural fields \cite{mildenhall2020nerf}, where each $d$-sized embedding $l_{t}^{k}$ represents at the same time the sample and its relative observation viewpoint.
At each time step $t$, we embed each image sample into the first layer of the columns, which is represented as the bottom layer $L_{t}^{0}$.

%At the embedding stage each input sample is divided in $n = h \times w$ patches of size $H_p\times W_p$ pixels, as shown in Fig. \ref{fig:teaser}. We then embed each patch at location $(h,w)$ into the bottom level $l_{t}^{0}$ of the corresponding column $C(h,w)$. In the embedding procedure, shown in Fig. \ref{fig:architecture}, we first rearrange the input data from $c(h\ H_p)(w\ W_p)$ to $(h\ w)(H_p\ W_p\ c)$, then we apply a linear layer, thus obtaining a vector of dimensions $(h\ w\ d)$.
%The second step is to perform the positional encoding \cite{mildenhall2020nerf} of each patch $n$ in position $(h\ w)$. 

\subsection{Hypercolumns}
\label{sec:hypercolumns}

Consecutive levels in time and space in a column $C(h,w)$ are connected by an auto-encoder. The auto-encoders are based on an MLP, which allows for model reduction \cite{hinton2015distilling} and faster training time. Each auto-encoder computes the top-down contribution of a level $l_{t-1}^{k+1}$ to the value of the level below at the next time step $l_{t}^{k}$ using a $N_{TD}(l_{t-1}^{k+1})$ top-down decoder. Similarly, each auto-encoder computes the bottom-up contribution of a level $l_{t-1}^{k-1}$ to the value of the level above at the next time step $l_{t}^{k}$ using a $N_{BU}(l_{t-1}^{k-1})$ bottom-up encoder.
$N_{TD}(l_{t-1}^{k+1})$ and $N_{BU}(l_{t-1}^{k-1})$ share a similar structure, but for the activation functions, as described in Fig. \ref{fig:architecture}(e). The top-down network uses GELU activation functions \cite{hendrycks2016gaussian}, while the bottom up network relies on sinusoidal activation functions \cite{sitzmann2020implicit,sopena1999neural,wong2002handwritten}. All the $N_{TD}(l_{t-1}^{k+1})$ connecting $L_{t-1}^{k+1}$ to layer $L_{t}^{k}$  share the same weights. The same is true for the $N_{BU}(l_{t-1}^{k-1})$ connecting $L_{t-1}^{k-1}$ to layer $L_{t}^{k}$.

\begin{figure*}
    \centering
    \includegraphics[width=0.96\textwidth]{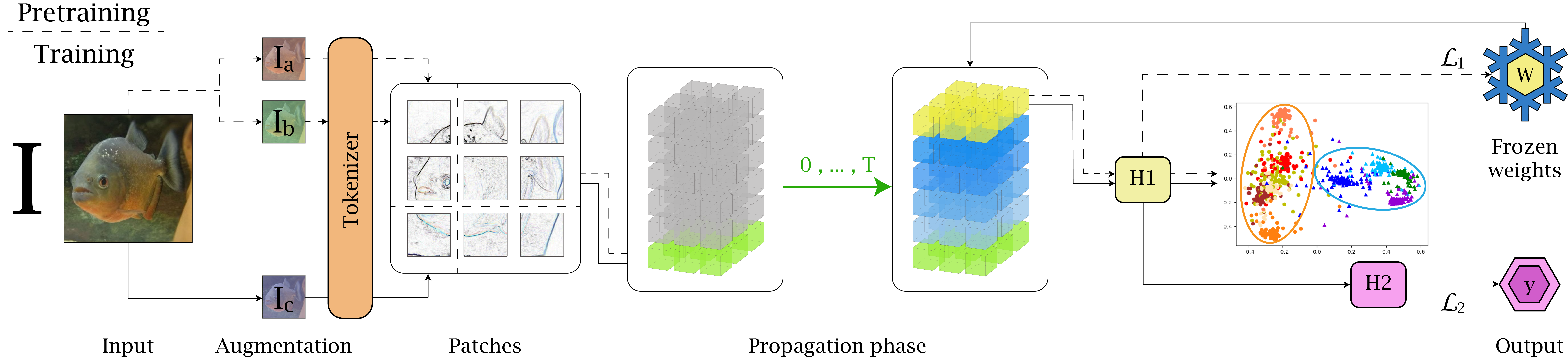}
    \caption{\textbf{Contrastive pre-training (dashed lines)} and \textbf{supervised training (continuous lines)} procedures. During the contrastive pre-training, two images $I_a$ and $I_b$ are produced by applying random data augmentation to the input image \textit{I}. Through the Tokenizer, we compute feature maps for both $I_a$ and $I_b$, which are then divided in patches and embedded into the bottom layer of the columns $L_{t}^{0}$. During the \textit{propagation phase}, the information is routed through the Agglomerator architecture to obtain the neural representation $L_{T}^{K}$ for each sample. We pre-train the network with the contrastive head $H1$ using a supervised contrastive loss $\mathcal{L}_1$, obtaining weights \textit{W}. During the supervised training, we first load the frozen weights $W$ in the network. Then, augmentation RandAugment \cite{cubuk2020randaugment} is applied on the input image $I$ to obtain $I_c$, which follows the same steps as the pre-training phase. The network, with the classification head $H2$, is trained for the classification task by minimizing the cross-entropy loss $\mathcal{L}_2$.}
    \label{fig:procedure}
\end{figure*}

\subsection{Routing}
\label{sec:routing}

The key element of our architecture is how the information is routed to obtain a representation of the input data where the part-whole hierarchies emerge.

Before computing the loss, we need to iteratively propagate each batch $N$ through the network, obtaining a deep representation of each image. This procedure, \textit{propagation phase}, encourages the network to reach \textit{consensus} between neighbor levels $l_{t}^{k} \in L_{t}^{k}$. Ideally, this means that all neighbor levels in the last layer $l_{t}^{K} \in L_{t}^{K}$ should have similar values, representing the same \textit{whole}; neighbor levels at bottom layers $l_{t}^{k} \in L_{t}^{k} | k \neq K$ should instead share the value among smaller groups, each group representing the same \textit{part}. Group of vectors that "agree" on a similar value have reached the \textit{consensus} on the image representation at that level, and they are called \textit{islands of agreement} \cite{hinton2021represent}. An example of such representation is shown in Fig. \ref{fig:teaser}(d). In capsules-based approaches \cite{sabour2017dynamic}, group of neurons are activated to represent the part-whole hierarchy with limited expressive power. Our $d$-dimensional layers $l_{t}^{k}$ provide a richer representation of the same hierarchy.

To obtain such representation, at time step $t=0$, we randomly initialise all the values $l_{0}^{k}$ and we embed a batch of $B$ samples into the bottom layer $L^0_0$. Once the values are initialized, we compute the attention $A(L_{t}^{k})$. Instead of the self-attention mechanism used in Transformers \cite{vaswani2017attention,dosovitskiy2020image,devlin2018bert}, a standard attention weighting is deployed as in \cite{xu2015show}. Each attention weight $\Omega_{n}$ is computed as

\vspace*{-7pt}
\begin{equation} 
\label{eq:attention_weights} 
\centering
    \begin{split}
     \Omega_{n} = \frac{e^{\beta \lambda_n \cdot l_{t}^{k}}}{\sum e^{\beta \lambda_n \cdot {N(\lambda_n)}}}
    \end{split}
\end{equation}
\vspace*{-7pt}

where $\lambda_n $ represents each possible level $l_{t}^{k}$ belonging to the same layer $L^k_t$ as $l^k_t$, $N(\lambda_n)$ is an indicator function which indexes all the neighbors levels of $\lambda_n $ belonging to the same layer $L_{t}^{k}$ and $\beta$ is a parameter that determines the sharpness of the attention.

At each time step $t \mid t \in \{1,\dots,T\}$, a batch with $B$ samples is fed to the bottom layer $L_t^0$ network as described in Sec. \ref{sec:embedding}. We compute the values $l_{t}^{k}$ as

% \begin{equation} 
% \label{eq:levels}
% \centering
%     %l(t) &= \sum_{N} BU(l-1(t-1)) + TD(l+1(t-1)) + l(t-1) + A(L(t-1))
%     l_{t}^{k} = \frac{ \omega_l l_{t-1}^{k} + \omega_{BU} N_{BU}(l_{t-1}^{k-1}) + \omega_{TD} N_{TD}(l_{t-1}^{k+1}) + \omega_A A(L_{t-1}^{k})}{ \#\ of\ contributions}
% \end{equation}

\begin{equation} 
\label{eq:levels}
\centering
    %l(t) &= \sum_{N} BU(l-1(t-1)) + TD(l+1(t-1)) + l(t-1) + A(L(t-1))
    \begin{matrix}
    l_{t}^{k} = avg(\omega_l l_{t-1}^{k},\omega_{BU} N_{BU}(l_{t-1}^{k-1}),\\
    \omega_{TD} N_{TD}(l_{t-1}^{k+1}),\omega_A A(L_{t-1}^{k}))
    \end{matrix}
\end{equation}

where $avg()$ indicates the arithmetical average, and $\omega_l,\omega_{BU},\omega_{TD},\omega_A$ are trainable weights. For layer $L^K_t$, contribution $N_{TD}(l_{t-1}^{k+1})$ is not included, as $L^{K+1}_t$ does not exist.
The \textit{propagation phase} takes $T$ time steps to reach the final representation of each image at each layer $L_{k}^{T}$.

\begin{figure*}[]
    \centering
    \begin{minipage}[b]{.55\textwidth}
    \centering
    \includegraphics[width=\textwidth]{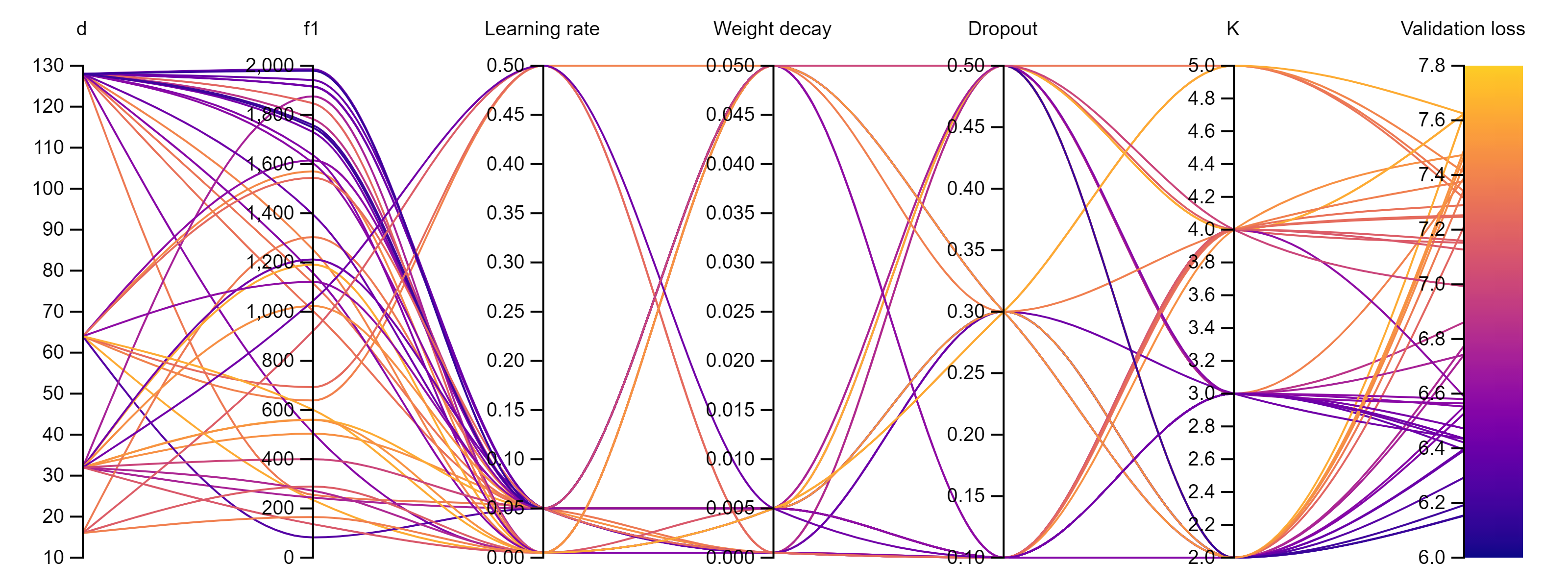}
    \caption{Hyper-parameters sweep. Each line represents a combination of parameters setup, with the darker lines representing the models achieving the lowest validation loss. Image obtained with \cite{wandb}.}
    \label{fig:sweep}
    \end{minipage}
    \qquad
    \begin{minipage}[b]{.33\textwidth}
    \centering
    
    % \includegraphics[width=\textwidth]{weights.png}
    % \caption{ Each line represents the variation of weights $\omega_l,\omega_{BU},\omega_{TD},\omega_A$ across epochs. Image obtained with \cite{wandb}.}
    % \label{fig:weights}

    \resizebox{\textwidth}{!}{%
    \begin{tabular}{@{}clc@{}}
    \toprule
    \multicolumn{1}{l}{} & Configuration            & Error \% (after 100 epochs) \\ \midrule
    I                    & Vanilla (proposed)                 & 12.8                        \\
    II                   & ReLU activation only    & 12.6                        \\
    III                  & Without attention       & 12.7                        \\
    IV                   & Linear columns layers         & 13.5                        \\
    V                    & Linear contrastive head & 15.8                        \\
    VI                   & Linear embedding        & 17.2                        \\ \bottomrule
    \end{tabular}%
    }
    \vspace{12px}
    \captionof{table}{Ablation study results of different Agglomerator configurations obtained on CIFAR-10 trained for 100 epochs. %Our network (I) performs similarly with (II) and (III). Both sinusoidal activations and shared attention in (I) are key to providing interpretable results, allowing islands of agreement to emerge.  Simplified versions using only a linear layer instead of column layers (IV), of the contrastive head (V), or of the linear embedding (VI) lead to a decrease in performance.
    }
    \label{tab:ablation_studies}
    
    \end{minipage}
\end{figure*}

\subsection{Training}
\label{sec:training}

The training procedure of our architecture is shown in Fig. \ref{fig:procedure}. It is divided in two steps: (i) a pre-training phase using a supervised contrastive loss function \cite{chen2020simple} and (ii) a training phase for the image classification using a Cross-Entropy loss. 

We first pre-train our network using an image-based contrastive loss \cite{chen2020simple}. Given a batch with $B$ samples, we duplicate each image $I$ to obtain pairs of samples $(I_a,I_b)$, for a total of $2B$ data points. We then apply data augmentation RandAugment \cite{cubuk2020randaugment} to both $(I_a,I_b)$. Both samples are fed to the network as described in Sec. \ref{sec:embedding}, and we perform the propagation phase in Sec. \ref{sec:routing} to obtain the representation at the last layer $L_{T}^{K}$.
Then we rearrange the $n$ levels $l_{T}^{K} \in L_{T}^{K}$ to obtain a vector of dimensions $n \times d$, given as input to the contrastive head $H1$, as described in Fig. \ref{fig:architecture}. At the output of the contrastive head, each sample is described by a feature vector of dimension $f1$. We take all the possible sample pairs $(I_a,I_b)$ from the batch and we compute the contrastive loss defined as:

\vspace*{-7pt}
\begin{equation}
\label{eq:contrastive_loss} 
\centering
    \begin{split}
    \mathcal{L}_1 &= ContrLoss(I_a,I_b) = - log \frac{e^{sim(I_a,I_b)}}{\sum^{2B}_{k=1} \mathcal{I}_{[k \neq a]}e^{sim(I_a,I_b)}}
    \end{split}
\end{equation}
\vspace*{-7pt}

where $sim(u,v) = u^T v/\norm{u}\norm{v}$ indicates the dot product between the normalised version of $u$ and $v$, $\mathcal{I}_{[k \neq a]}$ is a indicator function valued $0$ if $k$ and $a$ belong to the same class, and $1$ otherwise.

Once the network is pre-trained using the contrastive loss, the weights are frozen. We apply augmentation \cite{cubuk2020randaugment} to each sample $I_c$ in a batch of size $B$, which is then fed to the network for the \textit{propagation phase} to obtain for each sample the representation $L_{T}^{K}$. Then, the cross-entropy head $H2$ is added on top of the contrastive head $H1$. A linear layer resizes $f1$-dimensional features to dimension $f2$, which corresponds to the number of classes to be predicted for each dataset. The new layers are then trained using the cross-entropy function:

\vspace*{-7pt}
\begin{equation} 
\label{eq:ce_loss} 
\centering
    \begin{split}
       \mathcal{L}_2 = CE(y,\hat{y}) = - \frac{1}{f2}\sum^{\textrm{f2}}_{i=1} y_{i}\log({\hat{y}}_{i})
    \end{split}
\end{equation}
\vspace*{-7pt}

where ${y}$ is the label of a sample taken from the batch and $\hat{y}$ is the label to be predicted.

\section{Experiments}
\label{sec:experiments}

We perform our experiments on the following datasets:
    \par \textbf{SmallNorb (S-NORB)} \cite{lecun2004learning} is a dataset for 3D object recognition from shape. It consists of roughly 200000 images of size $96 \times 96$ pixels of 5 classes of toys.
    \par \textbf{MNIST} \cite{lecun1998gradient} and \textbf{FashionMNIST} \cite{xiao2017/online}, consist of 60000 training images and 10000 test images of grayscale handwritten digits and Zalando's articles of size $28 \times 28$ pixels.
    \par \textbf{CIFAR-10} and \textbf{CIFAR-100} \cite{krizhevsky2009learning} both consist of 50000 training images and 10000 test images of size $32 \times 32$ pixels, with 10 and 100 classes, respectively.

Our network is trained in an end-to-end fashion using PyTorch Lightning on a single NVIDIA GeForce RTX 3090. Input images for each dataset are normalized using each standard dataset's normalization. We train our network on each dataset's native resolution, except for SmallNorb, which is resized to $32\times 32$ pixels, following the standard procedure as in \cite{ribeiro2020capsule,hinton2018matrix}. The Tokenizer embedding creates $n = H/4 \times W/4$ patches represented by $n$ $d$-dimensional vectors, where $H$ and $W$ are the pixels dimension of the input image. Thus the corresponding number of columns is $8 \times 8$ for CIFAR-10, CIFAR-100, and SmallNorb, and $7 \times 7$ for MNIST FashionMNIST. During the pre-training, we deploy the following hyper-parameters: $300$ epochs, cyclic learning rate \cite{smith2017cyclical} in the range $[0.002, 0.05]$, batch size $B=1024$, levels embedding $d = 128$, number of levels $K=3$, number of iterations $T=2K=6$, dropout value $0.3$, contrastive features dimension $f1=512$, and weight decay $5e^{-4}$. During the training phase, we resume the network training with the same hyper-parameters, $f2$ being the number of classes corresponding to each dataset.
% \begin{itemize}
%     \item image size, canali, numero classi, numero di patch cambia mnist e fascion 7, cifar 8, small ridimensionato 48 x48 seguendo procedur paper \cite{ribeiro2020capsule,hinton2018matrix}, quindi 32x32, quindi 8x8, quindi patch size 1
%     \item numero epoche 300 per tutti, 
%     \item learning rate $0.002=0.05$ scheduler ciclico, dove finisce il pre-train inizia quello del train
%     \item batch size $N=1024$
%     \item patch size $1 feature-pixel$
%     \item $d=128$ con piu memoria si puo modificare
%     \item levels $3$
%     \item dropout $0.3$
%     \item contrastive dimension $f1 = 512$
%     \item ce dimension $f2 = num_class$
%     \item weight decay $5e^{-4}$
% \end{itemize}

% Our network is trained in an end-to-end fashion using Pytorch Lightning. Input images are normalized in the interval $[0,1]$ with a resolution of 256x256 pixels for depth images and 256x256 pixels for RGB ones. We do not perform any augmentations on the input datasets. The batch size is set to 128 for ITOP and 128 for PanopTOP31K. We initialize the weights with the Xavier initialization \cite{glorot2010understanding}. The learning rate is set to $1e^{-5}$, the weight decay is set to 0, and Adam is the optimizer of choice. We train our network for 20 epochs on the ITOP dataset and 15 epochs on PanopTOP31K.

% Tabella con parametri per ogni dataset?

\begin{table*}[]
\centering
\resizebox{.85\textwidth}{!}{%
\begin{tabular}{cccccccccc}
\hline
\multirow{3}{*}{\textbf{Method}} & \multirow{3}{*}{\textbf{Ref}}                       & \multirow{3}{*}{\textbf{Backbone}} & \multicolumn{5}{c}{\textbf{Error \%}}                                                                                                                                                                                                                                         & \multirow{3}{*}{\textbf{\begin{tabular}[c]{@{}c@{}}\# of \\ params\\  (Millions)\end{tabular}}} & \multirow{3}{*}{\textbf{\begin{tabular}[c]{@{}c@{}}Training \\ Arch.\end{tabular}}} \\ \cline{4-8}
                                 &                                                     &                                    & \multicolumn{1}{l}{\multirow{2}{*}{\textbf{S-Norb}}} & \multicolumn{1}{l}{\multirow{2}{*}{\textbf{MNIST}}} & \multicolumn{1}{l}{\multirow{2}{*}{\textbf{F-MNIST}}} & \multicolumn{1}{l}{\multirow{2}{*}{\textbf{C-10}}} & \multicolumn{1}{l}{\multirow{2}{*}{\textbf{C-100}}} &                                                                                                 &                                                                                     \\
                                 &                                                     &                                    & \multicolumn{1}{l}{}                                 & \multicolumn{1}{l}{}                                & \multicolumn{1}{l}{}                                  & \multicolumn{1}{l}{}                               & \multicolumn{1}{l}{}                                &                                                                                                 &                                                                                     \\ \hline
E-CapsNet                        & \cite{mazzia2021efficient}                          & \multirow{4}{*}{Caps}              & 2.54                                                 & 0.26                                                & -                                                     & -                                                  & -                                                   & 0.2                                                                                             & GPU                                                                                 \\
CapsNet                          & \cite{sabour2017dynamic,mukhometzianov2018capsnet}  &                                    & 2.70                                                 & 0.25                                                & 6.38                                                  & 10.6                                               & 82.00                                               & 6.8                                                                                             & GPU                                                                                 \\
Matrix-CapsNet                   & \cite{hinton2018matrix}                             &                                    & 1.40                                                 & 0.44                                                & 6.14                                                  & 11.9                                               & -                                                   & 0.3                                                                                             & GPU                                                                                 \\
Capsule VB                       & \cite{ribeiro2020capsule}                           &                                    & 1.60                                                 & 0.30                                                & 5.20                                                  & 11.2                                               & -                                                   & 0.2                                                                                             & GPU                                                                                 \\ \hline
ResNet-110                       & \cite{he2016deep,huang2016deep,assunccao2019denser} & \multirow{2}{*}{Conv}              & -                                                    & 2.10                                                & 5.10                                                  & 6.41*                                              & 27.76*                                              & 1.7                                                                                             & GPU                                                                                 \\
VGG                              & \cite{simonyan2014very,assunccao2019denser}         &                                    & -                                                    & 0.32                                                & 6.50                                                  & 7.74*                                              & 28.05*                                              & 20                                                                                              & GPU                                                                                 \\ \hline
ViT-L/16                         & \cite{dosovitskiy2020image}                         & Transf                             & -                                                    & -                                                   & -                                                     & 0.85*                                              & 6.75*                                               & 632                                                                                             & TPU                                                                                 \\ \hline
ConvMLP-L                        & \cite{li2021convmlp}                                & Conv/MLP                           & -                                                    & -                                                   & -                                                     & 1.40*                                              & 11.40*                                              & 43                                                                                              & TPU                                                                                 \\
MLP-Mixer-L/16                   & \cite{tolstikhin2021mlp}                            & MLP                                & -                                                    & -                                                   & -                                                     & 1.66*                                              & -                                                   & 207                                                                                             & TPU                                                                                 \\ \hline
\textbf{Ours}                    &                                                     & Conv/MLP/Caps                      & 0.01                                                 & 0.30                                                 & 7.43                                                  & \multicolumn{1}{r}{11.15}                          & \multicolumn{1}{r}{40.97}                           & 72                                                                                              & GPU                                                                                 \\ \hline
\end{tabular}%
}
\caption{Error percentages on the Top-1 accuracy results on datasets SmallNorb (S-Norb), MNIST, FashionMNIST (F-MNIST), CIFAR-10 (C-10), and CIFAR-100 (C-100). The $*$ notation indicates results obtained with networks pre-trained on ImageNet.}
\label{tab:results}
\end{table*}

\begin{figure*}
    \centering
    \includegraphics[width=.93\textwidth]{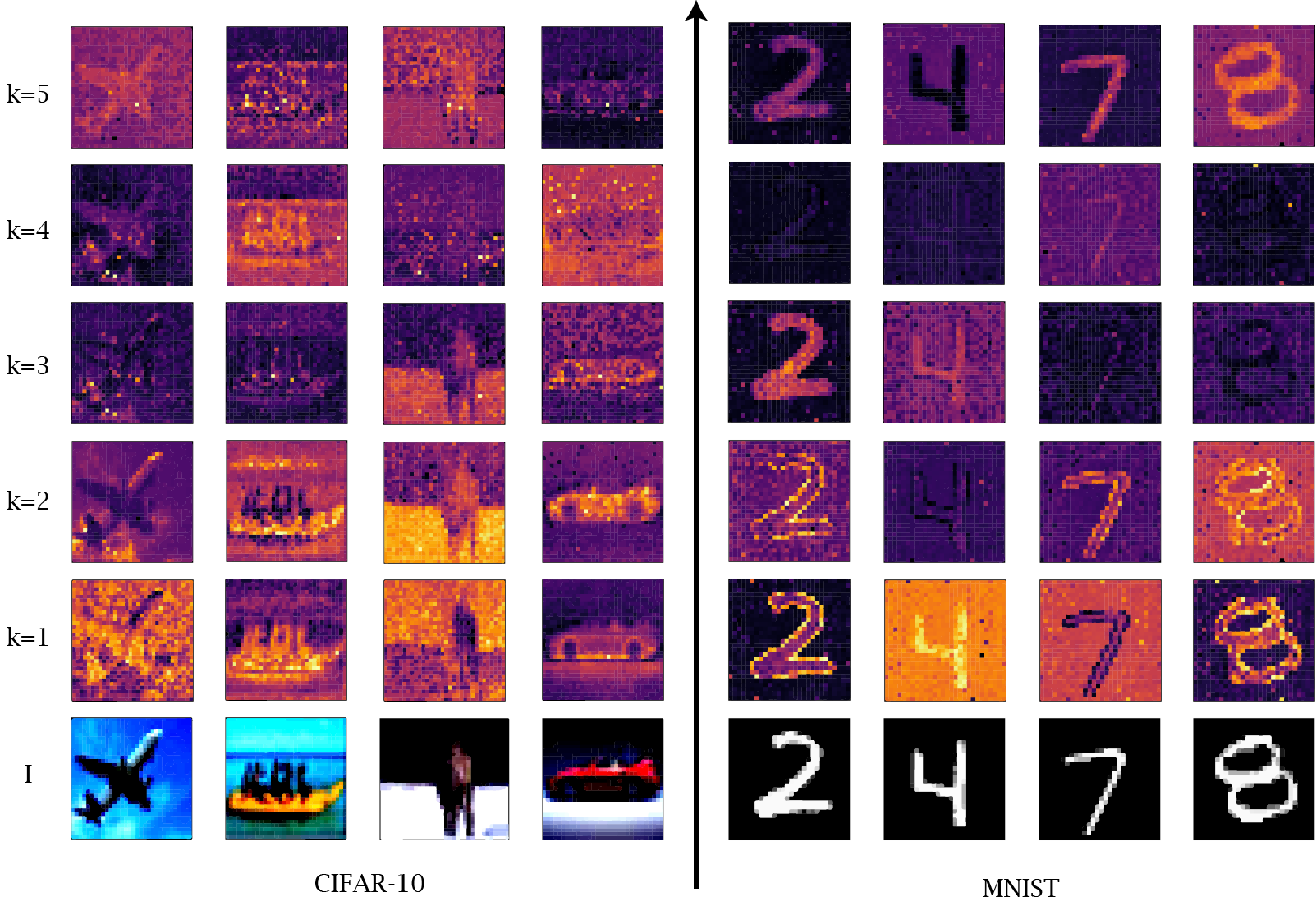}
    \caption{Vectorial representation of emerging islands of agreement at different $K$ levels of sample from MNIST and CIFAR-10 datasets. We show the vectors of agreement for each patch at each level $k$ after 100 epochs of contrastive pre-training. At level $k=1$, the network acts similarly to a feature extractor, where each cell represent a spatial feature with little agreement between neighbors. At intermediate levels, $k=2,3,4$ neighbor cells reach agreement on specific parts of the image, creating different island for different part of the plane. At the last level $k=5$, two island emerge, agreeing on the representation of the object and of the background. Since we are training the network for the classification task, the distance between the color of the two island is small since all the parts of the image tend to agree that the image represents the same whole.}
    \label{fig:islands_full}
% \end{figure}
% \begin{figure}[!h]
    % \centering
    % \includegraphics[width=\textwidth]{islands_detail.png}
    % \caption{Islands of agreement detail}
    % \label{fig:islands_detail}
\end{figure*}

\section{Quantitative results}
\label{sec:quantitative_results}

We report the quantitative results for each dataset in Tab. \ref{tab:results}. Capsule-based models \cite{hinton2018matrix,sabour2017dynamic,mazzia2021efficient,mukhometzianov2018capsnet,ribeiro2020capsule} can achieve good performances on simple datasets (SmallNorb, MNIST, and FashionMNIST), but they fail to generalize to datasets with a higher number of classes (CIFAR-100). Convolutional-based models \cite{he2016deep,assunccao2019denser,huang2016deep,simonyan2014very} can generalize to different datasets, at the expense of weak model interpretability, mainly due to the max-pooling operation.
Transformer-based \cite{dosovitskiy2020image} and MLP-based methods \cite{li2021convmlp,tolstikhin2021mlp} are able to achieve the best performances on more complex datasets, but they do not provide tests for smaller datasets. However, to achieve such levels of accuracy they rely on long pretraining (thousands of TPU days) on expensive computational architectures, implementing data augmentation on ImageNet \cite{krizhevsky2012imagenet} or the JFT-300M \cite{sun2017revisiting} dataset, not available publicly. As can be seen, our method performs on par with capsule-based methods on simpler datasets, while achieving better generalization on more complex ones. In addition, our method has fewer parameters than most transformer-based and MLP-based methods, and it can be trained in less time on a much smaller architecture.

% parametri con mlp, prestazioni con capsule e in generale con modelli non trainati su fdataset grandi come il demonio, 

% 1 da un parte emtodi compresi capsule che funizonano bene su dataset semplici (pochi param)
% 2transforme e metodi giganti che che funzionao bene su dataset grossi, ma hanno bisognio di pretraining spinto senno fanno piu schifo delle resnet (bordello di paramtri)
% 3 noi facciamo decentemente su dataset piccoli e medi con meno training data e molti meno parametri e molta piu interpretabilita.

% struttura di pretaining o reti piu grandi, noi molot efficenti

\begin{figure*}[]
\centering
        \begin{subfigure}[b]{0.27\textwidth}
                \includegraphics[width=\linewidth]{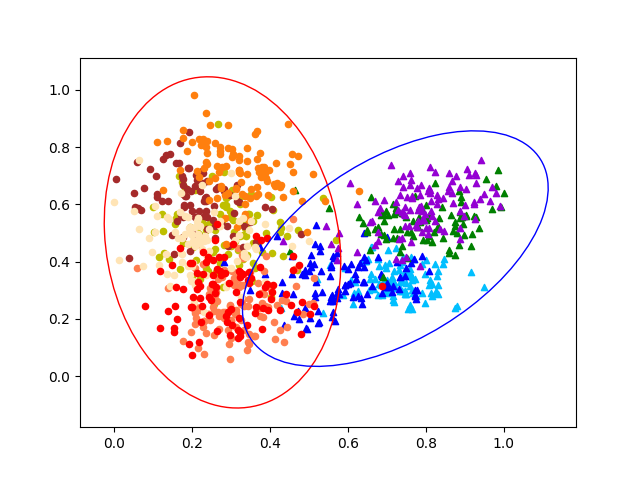}
                \caption{ResNet-110 \cite{he2016deep} \\\textbf{O=12\%}}
        \end{subfigure}%
        \begin{subfigure}[b]{0.27\textwidth}
                \includegraphics[width=\linewidth]{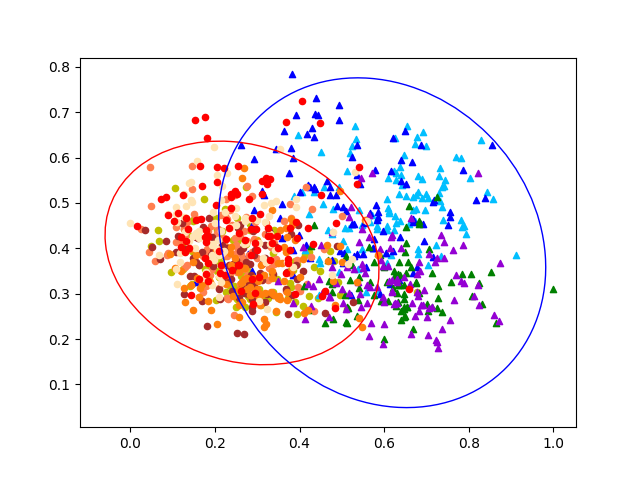}
                \caption{ViT-L/16 \cite{dosovitskiy2020image} \\\textbf{O=24\%}}
        \end{subfigure}%
        \begin{subfigure}[b]{0.27\textwidth}
                \includegraphics[width=\linewidth]{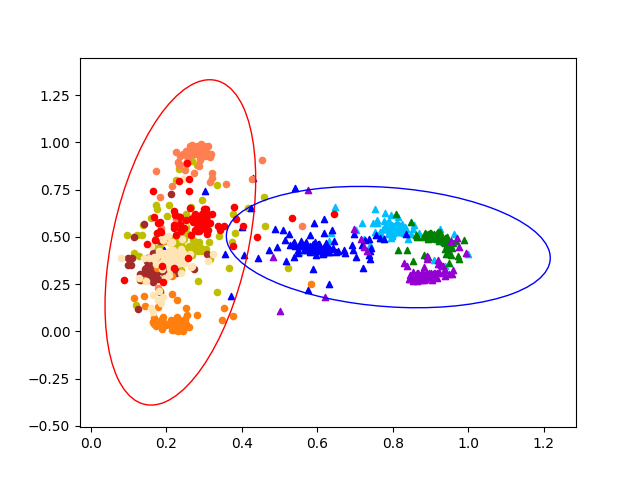}
                \caption{Ours \\\textbf{O=2\%}}
        \end{subfigure}%
        \\
        \begin{subfigure}[b]{0.27\textwidth}
                \includegraphics[width=\linewidth]{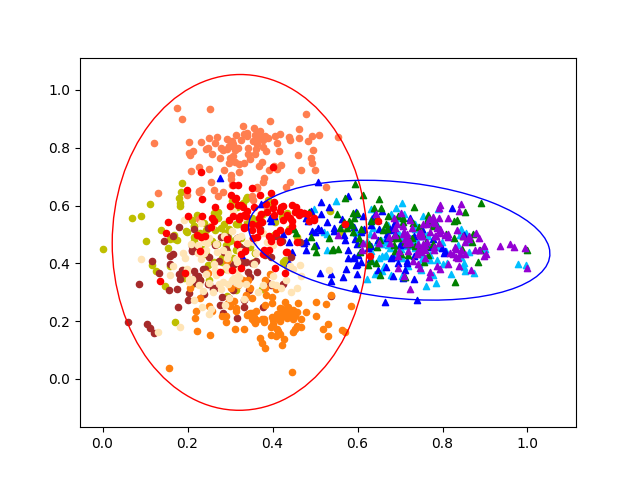}
                \caption{ConvMLP-L \cite{li2021convmlp} \\\textbf{O=12\%}}
        \end{subfigure}%
        \begin{subfigure}[b]{0.27\textwidth}
                \includegraphics[width=\linewidth]{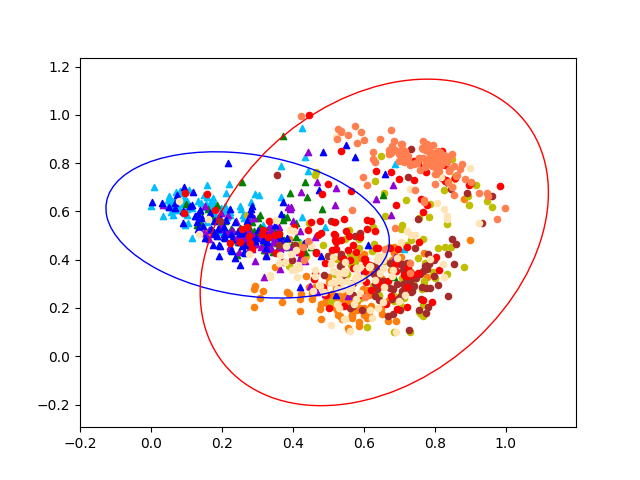}
                \caption{Matrix-CapsNet \cite{hinton2018matrix} \\\textbf{O=20\%}}
        \end{subfigure}%
        \begin{subfigure}[b]{0.27\textwidth}
                \includegraphics[width=\linewidth]{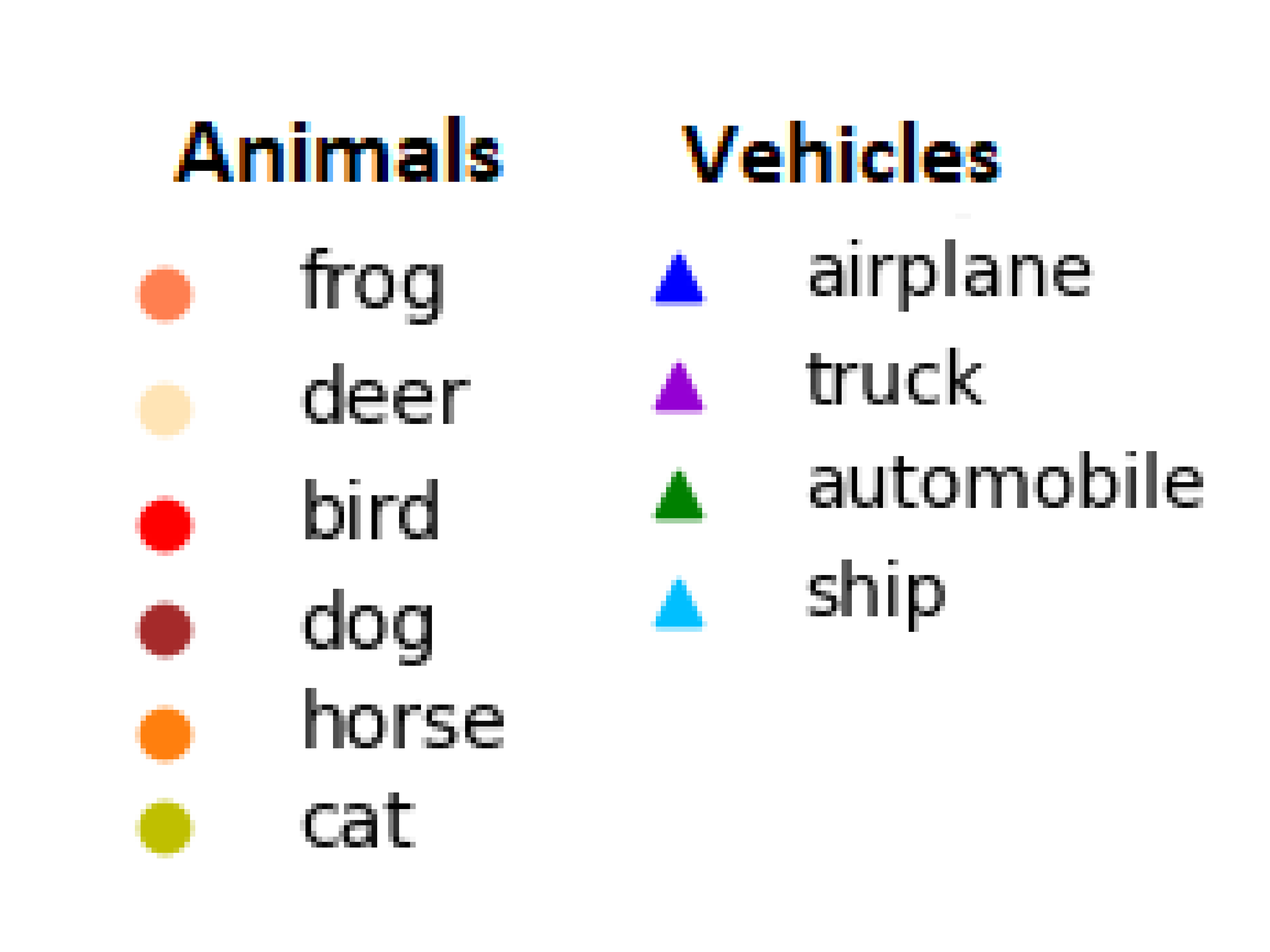}
                \caption{Legend}
        \end{subfigure}%
        \caption{2D representation of the latent space for multiple methods trained only on the CIFAR-10 dataset obtained using Principal Component Analysis (PCA) \cite{wold1987principal}. The PCA provides a deterministic change of base for the data from a multidimensional space into a 2D space. The legend (f) displays the classes, which are divided between super-classes \textit{Vehicles} and \textit{Animals} following the WordNet hierarchy \cite{miller1995wordnet}. The different methods (a,b,c,d,e) are all able to cluster the samples between the two super-classes. However, while (a,b,e) display a latent space where classes are close to each other, the two MLP-based methods (c,d) are able to provide a clearer separation between the super-classes. Both methods show conceptual-semantically close samples on the edge of each superclass, such as airplanes and birds. Inside each superclass, semantically close samples are represented contiguously, such as deers and horses, or cars and trucks. Our method (c) provides better inter-class and intra-class separability. The overlap percentage $O$ is reported for each method. The overlap area is the area where a mistake with a higher hierarchical severity \cite{bertinetto2020making} has a higher probability to occur.
        }
        \label{fig:latent}
\end{figure*}

\textbf{Ablation study.} We analyze the contribution of the different components of our architecture evaluating their influence on the validation loss after $50$ epochs. The considered parameters, in descending order of correlation with the validation loss value are: the embedding dimension $d$, the contrastive feature vector $f1$, learning rate, weight decay, dropout, and the number of levels $K$. The results are reported in Fig. \ref{fig:sweep}. We perform $50$ different training on CIFAR-10 with different combinations of parameters.

In Tab. \ref{tab:ablation_studies} we show how our network configuration (I) performs similarly with (II) and (III). Both sinusoidal activations and shared attention in (I) are key to providing interpretable results, allowing islands of \textit{agreement to emerge}.  Simplified versions using only a linear layer instead of column layers (IV), of the contrastive head (V), or of the linear embedding (VI) lead to a decrease in performance.

% In Fig. \ref{fig:weights}, we show how the contributions to the values of levels $l^k_t$ in Eq. \ref{eq:levels} are weighted over the first 100 epochs by observing how the trainable weights $\omega_l,\omega_{BU},\omega_{TD},\omega_A$ change. The attention contribution $\omega_A$ and the previous level one $\omega_l$ decrease over time, with $\omega_l$ reaching a plateau.
% After 15 epochs of weights self-calibration, the top-down contribution $\omega_{TD}$ increases over the epochs, as the network can infer useful information from predicted "wholes" at the upper levels.
% The bottom-up contribution $\omega_{BU}$ decreases over the epochs since the bottom-up networks have learned to map efficiently map the input into the embeddings, especially at lower levels.

\section{Qualitative results: interpretability}
\label{sec:qualitative_results}

Our method provides interpretability of the \textit{relationships learned by the model} by explicitly modeling the part-whole hierarchy, and of the \textit{relationships contained in data} through the hierarchical organization of the feature space.

\textbf{Island of agreement as a representation of multi-level part-whole hierarchy.} During the \textit{propagation phase}, neighbor levels on the same layer $L^k_t$ are encouraged to reach a \textit{consensus} by forming \textit{islands of agreement}. The \textit{islands of agreement} represent the part-whole hierarchies at different levels. In Fig. \ref{fig:islands_full}, we provide a few examples of the islands of agreement obtained on MNIST and CIFAR-10 trained with $K=5$ levels. Each arrow represents the value of a level $l_k^t$ at location $(h,w)$, reduced from $d$-dimensional to 2D using a linear layer. As $k$ for $L^k_t$ increases, neighbor $l_k^t \in L^k_t$ tend to agree on a common representation of the \textit{whole} represented in the image sample. At lower levels, smaller islands emerge, each representing a part of the \textit{whole}. Samples of MNIST present fewer changes in the islands across levels because the data is much simpler, indicating that fewer levels in the hierarchy can be sufficient to obtain similar results. Our Agglomerator is thus able to represent a patch differently at different levels of abstraction. At the same level, spatially adjacent patches take the same value, agreeing on the representation of parts and wholes.

\textbf{Latent space organization as the representation of conceptual-semantic relationship in data.} Recent networks aim at maximizing inter-class distances and minimizing intra-class distances between samples in the latent space. While the accuracy is high, they provide little interpretability in their data representation. As a result, mistakes are less likely to happen, but the mistake severity, defined as the distance between two classes in WordNet lexical hierarchy \cite{miller1995wordnet}, does not decrease \cite{bertinetto2020making}. As shown in Fig. \ref{fig:latent}, our network semantically organizes the input data resembling the human lexical hierarchy.

\section{Limitations}

Our method introduces new types of hyper-parameters in the network structure, such as embedding dimensions, number of levels, and size of patches, which need to be tuned. We believe a better parameters setting can be found for all the datasets, increasing accuracy while still retaining interpretability. Moreover, a higher number of parameters generally causes architectures to be more prone to over-fitting and more difficult to train.
To improve the accuracy of our network, we would need a pre-training on large datasets (e.g., on ImageNet), which requires large computational resources to be performed in a reasonable time frame. While hoping that powerful TPU architectures become publicly available in the future, we are currently investigating efficient pre-training strategies for our network.

\vspace*{-5px}
\section{Conclusion}

We presented Agglomerator, a method that makes a step forward towards representing interpretable part-whole hierarchies and conceptual-semantic relationships in neural networks. We believe that interpretable networks are key to the success of artificial intelligence and deep learning. With this work, we intend to promote a preliminary implementation and the corresponding results on the image classification task, and we hope to inspire other researchers to adjust our solution to solve more complex and diverse tasks.

%%%%%%%%% REFERENCES
\clearpage

{\small
\bibliographystyle{ieee_fullname}
\bibliography{egbib}

\begin{thebibliography}{10}\itemsep=-1pt

\bibitem{albarracin2000cognitive}
Dolores Albarracin and Robert~S Wyer~Jr.
\newblock The cognitive impact of past behavior: influences on beliefs,
  attitudes, and future behavioral decisions.
\newblock {\em Journal of personality and social psychology}, 79(1):5, 2000.

\bibitem{assunccao2019denser}
Filipe Assun{\c{c}}{\~a}o, Nuno Louren{\c{c}}o, Penousal Machado, and
  Bernardete Ribeiro.
\newblock Denser: deep evolutionary network structured representation.
\newblock {\em Genetic Programming and Evolvable Machines}, 20(1):5--35, 2019.

\bibitem{bear2020learning}
Daniel~M Bear, Chaofei Fan, Damian Mrowca, Yunzhu Li, Seth Alter, Aran Nayebi,
  Jeremy Schwartz, Li Fei-Fei, Jiajun Wu, Joshua~B Tenenbaum, et~al.
\newblock Learning physical graph representations from visual scenes.
\newblock {\em arXiv preprint arXiv:2006.12373}, 2020.

\bibitem{bertinetto2020making}
Luca Bertinetto, Romain Mueller, Konstantinos Tertikas, Sina Samangooei, and
  Nicholas~A Lord.
\newblock Making better mistakes: Leveraging class hierarchies with deep
  networks.
\newblock In {\em Proceedings of the IEEE/CVF Conference on Computer Vision and
  Pattern Recognition}, pages 12506--12515, 2020.

\bibitem{biederman1987recognition}
Irving Biederman.
\newblock Recognition-by-components: a theory of human image understanding.
\newblock {\em Psychological review}, 94(2):115, 1987.

\bibitem{wandb}
Lukas Biewald.
\newblock Experiment tracking with weights and biases, 2020.
\newblock Software available from wandb.com.

\bibitem{chen2020simple}
Ting Chen, Simon Kornblith, Mohammad Norouzi, and Geoffrey Hinton.
\newblock A simple framework for contrastive learning of visual
  representations.
\newblock In {\em International conference on machine learning}, pages
  1597--1607. PMLR, 2020.

\bibitem{cubuk2020randaugment}
Ekin~D Cubuk, Barret Zoph, Jonathon Shlens, and Quoc~V Le.
\newblock Randaugment: Practical automated data augmentation with a reduced
  search space.
\newblock In {\em Proceedings of the IEEE/CVF Conference on Computer Vision and
  Pattern Recognition Workshops}, pages 702--703, 2020.

\bibitem{deng2020generative}
Fei Deng, Zhuo Zhi, Donghun Lee, and Sungjin Ahn.
\newblock Generative scene graph networks.
\newblock In {\em International Conference on Learning Representations}, 2020.

\bibitem{devlin2018bert}
Jacob Devlin, Ming-Wei Chang, Kenton Lee, and Kristina Toutanova.
\newblock Bert: Pre-training of deep bidirectional transformers for language
  understanding.
\newblock {\em arXiv preprint arXiv:1810.04805}, 2018.

\bibitem{doshi2017towards}
Finale Doshi-Velez and Been Kim.
\newblock Towards a rigorous science of interpretable machine learning.
\newblock {\em arXiv preprint arXiv:1702.08608}, 2017.

\bibitem{dosovitskiy2020image}
Alexey Dosovitskiy, Lucas Beyer, Alexander Kolesnikov, Dirk Weissenborn,
  Xiaohua Zhai, Thomas Unterthiner, Mostafa Dehghani, Matthias Minderer, Georg
  Heigold, Sylvain Gelly, et~al.
\newblock An image is worth 16x16 words: Transformers for image recognition at
  scale.
\newblock {\em arXiv preprint arXiv:2010.11929}, 2020.

\bibitem{grigorescu2020survey}
Sorin Grigorescu, Bogdan Trasnea, Tiberiu Cocias, and Gigel Macesanu.
\newblock A survey of deep learning techniques for autonomous driving.
\newblock {\em Journal of Field Robotics}, 37(3):362--386, 2020.

\bibitem{hawkins2021thousand}
Jeff Hawkins.
\newblock A thousand brains: A new theory of intelligence, 2021.

\bibitem{hawkins2017theory}
Jeff Hawkins, Subutai Ahmad, and Yuwei Cui.
\newblock A theory of how columns in the neocortex enable learning the
  structure of the world.
\newblock {\em Frontiers in neural circuits}, 11:81, 2017.

\bibitem{he2016deep}
Kaiming He, Xiangyu Zhang, Shaoqing Ren, and Jian Sun.
\newblock Deep residual learning for image recognition.
\newblock In {\em Proceedings of the IEEE conference on computer vision and
  pattern recognition}, pages 770--778, 2016.

\bibitem{hendrycks2016gaussian}
Dan Hendrycks and Kevin Gimpel.
\newblock Gaussian error linear units (gelus).
\newblock {\em arXiv preprint arXiv:1606.08415}, 2016.

\bibitem{hinton2021represent}
Geoffrey Hinton.
\newblock How to represent part-whole hierarchies in a neural network.
\newblock {\em arXiv preprint arXiv:2102.12627}, 2021.

\bibitem{hinton2015distilling}
Geoffrey Hinton, Oriol Vinyals, and Jeff Dean.
\newblock Distilling the knowledge in a neural network.
\newblock {\em arXiv preprint arXiv:1503.02531}, 2015.

\bibitem{hinton1990mapping}
Geoffrey~E Hinton.
\newblock Mapping part-whole hierarchies into connectionist networks.
\newblock {\em Artificial Intelligence}, 46(1-2):47--75, 1990.

\bibitem{hinton2018matrix}
Geoffrey~E Hinton, Sara Sabour, and Nicholas Frosst.
\newblock Matrix capsules with em routing.
\newblock In {\em International conference on learning representations}, 2018.

\bibitem{hole2021thousand}
Kjell~J{\o}rgen Hole and Subutai Ahmad.
\newblock A thousand brains: toward biologically constrained ai.
\newblock {\em SN Applied Sciences}, 3(8):1--14, 2021.

\bibitem{hong2021ptr}
Yining Hong, Li Yi, Josh Tenenbaum, Antonio Torralba, and Chuang Gan.
\newblock Ptr: A benchmark for part-based conceptual, relational, and physical
  reasoning.
\newblock {\em Advances in Neural Information Processing Systems}, 34, 2021.

\bibitem{huang2016deep}
Gao Huang, Yu Sun, Zhuang Liu, Daniel Sedra, and Kilian~Q Weinberger.
\newblock Deep networks with stochastic depth.
\newblock In {\em European conference on computer vision}, pages 646--661.
  Springer, 2016.

\bibitem{khan2021transformers}
Salman Khan, Muzammal Naseer, Munawar Hayat, Syed~Waqas Zamir, Fahad~Shahbaz
  Khan, and Mubarak Shah.
\newblock Transformers in vision: A survey.
\newblock {\em arXiv preprint arXiv:2101.01169}, 2021.

\bibitem{kosiorek2019stacked}
Adam~R Kosiorek, Sara Sabour, Yee~Whye Teh, and Geoffrey~E Hinton.
\newblock Stacked capsule autoencoders.
\newblock {\em arXiv preprint arXiv:1906.06818}, 2019.

\bibitem{krizhevsky2009learning}
Alex Krizhevsky, Geoffrey Hinton, et~al.
\newblock Learning multiple layers of features from tiny images.
\newblock 2009.

\bibitem{krizhevsky2012imagenet}
Alex Krizhevsky, Ilya Sutskever, and Geoffrey~E Hinton.
\newblock Imagenet classification with deep convolutional neural networks.
\newblock {\em Advances in neural information processing systems},
  25:1097--1105, 2012.

\bibitem{lecun2015deep}
Yann LeCun, Yoshua Bengio, and Geoffrey Hinton.
\newblock Deep learning.
\newblock {\em nature}, 521(7553):436--444, 2015.

\bibitem{lecun1998gradient}
Yann LeCun, L{\'e}on Bottou, Yoshua Bengio, and Patrick Haffner.
\newblock Gradient-based learning applied to document recognition.
\newblock {\em Proceedings of the IEEE}, 86(11):2278--2324, 1998.

\bibitem{lecun2004learning}
Yann LeCun, Fu~Jie Huang, and Leon Bottou.
\newblock Learning methods for generic object recognition with invariance to
  pose and lighting.
\newblock In {\em Proceedings of the 2004 IEEE Computer Society Conference on
  Computer Vision and Pattern Recognition, 2004. CVPR 2004.}, volume~2, pages
  II--104. IEEE, 2004.

\bibitem{lee2019set}
Juho Lee, Yoonho Lee, Jungtaek Kim, Adam Kosiorek, Seungjin Choi, and Yee~Whye
  Teh.
\newblock Set transformer: A framework for attention-based
  permutation-invariant neural networks.
\newblock In {\em International Conference on Machine Learning}, pages
  3744--3753. PMLR, 2019.

\bibitem{li2021convmlp}
Jiachen Li, Ali Hassani, Steven Walton, and Humphrey Shi.
\newblock Convmlp: Hierarchical convolutional mlps for vision.
\newblock {\em arXiv preprint arXiv:2109.04454}, 2021.

\bibitem{linardatos2021explainable}
Pantelis Linardatos, Vasilis Papastefanopoulos, and Sotiris Kotsiantis.
\newblock Explainable ai: A review of machine learning interpretability
  methods.
\newblock {\em Entropy}, 23(1):18, 2021.

\bibitem{liu2021swin}
Ze Liu, Yutong Lin, Yue Cao, Han Hu, Yixuan Wei, Zheng Zhang, Stephen Lin, and
  Baining Guo.
\newblock Swin transformer: Hierarchical vision transformer using shifted
  windows.
\newblock {\em arXiv preprint arXiv:2103.14030}, 2021.

\bibitem{mazzia2021efficient}
Vittorio Mazzia, Francesco Salvetti, and Marcello Chiaberge.
\newblock Efficient-capsnet: Capsule network with self-attention routing.
\newblock {\em arXiv preprint arXiv:2101.12491}, 2021.

\bibitem{mildenhall2020nerf}
Ben Mildenhall, Pratul~P Srinivasan, Matthew Tancik, Jonathan~T Barron, Ravi
  Ramamoorthi, and Ren Ng.
\newblock Nerf: Representing scenes as neural radiance fields for view
  synthesis.
\newblock In {\em European conference on computer vision}, pages 405--421.
  Springer, 2020.

\bibitem{miller1995wordnet}
George~A Miller.
\newblock Wordnet: a lexical database for english.
\newblock {\em Communications of the ACM}, 38(11):39--41, 1995.

\bibitem{miller2019explanation}
Tim Miller.
\newblock Explanation in artificial intelligence: Insights from the social
  sciences.
\newblock {\em Artificial intelligence}, 267:1--38, 2019.

\bibitem{miotto2018deep}
Riccardo Miotto, Fei Wang, Shuang Wang, Xiaoqian Jiang, and Joel~T Dudley.
\newblock Deep learning for healthcare: review, opportunities and challenges.
\newblock {\em Briefings in bioinformatics}, 19(6):1236--1246, 2018.

\bibitem{mukhometzianov2018capsnet}
Rinat Mukhometzianov and Juan Carrillo.
\newblock Capsnet comparative performance evaluation for image classification.
\newblock {\em arXiv preprint arXiv:1805.11195}, 2018.

\bibitem{murdoch2019definitions}
W~James Murdoch, Chandan Singh, Karl Kumbier, Reza Abbasi-Asl, and Bin Yu.
\newblock Definitions, methods, and applications in interpretable machine
  learning.
\newblock {\em Proceedings of the National Academy of Sciences},
  116(44):22071--22080, 2019.

\bibitem{ribeiro2020capsule}
Fabio De~Sousa Ribeiro, Georgios Leontidis, and Stefanos Kollias.
\newblock Capsule routing via variational bayes.
\newblock In {\em Proceedings of the AAAI Conference on Artificial
  Intelligence}, volume~34, pages 3749--3756, 2020.

\bibitem{sabour2017dynamic}
Sara Sabour, Nicholas Frosst, and Geoffrey~E Hinton.
\newblock Dynamic routing between capsules.
\newblock {\em arXiv preprint arXiv:1710.09829}, 2017.

\bibitem{sezer2020financial}
Omer~Berat Sezer, Mehmet~Ugur Gudelek, and Ahmet~Murat Ozbayoglu.
\newblock Financial time series forecasting with deep learning: A systematic
  literature review: 2005--2019.
\newblock {\em Applied Soft Computing}, 90:106181, 2020.

\bibitem{simonyan2014very}
Karen Simonyan and Andrew Zisserman.
\newblock Very deep convolutional networks for large-scale image recognition.
\newblock {\em arXiv preprint arXiv:1409.1556}, 2014.

\bibitem{sitzmann2020implicit}
Vincent Sitzmann, Julien Martel, Alexander Bergman, David Lindell, and Gordon
  Wetzstein.
\newblock Implicit neural representations with periodic activation functions.
\newblock {\em Advances in Neural Information Processing Systems}, 33, 2020.

\bibitem{sitzmann2019scene}
Vincent Sitzmann, Michael Zollh{\"o}fer, and Gordon Wetzstein.
\newblock Scene representation networks: Continuous 3d-structure-aware neural
  scene representations.
\newblock {\em arXiv preprint arXiv:1906.01618}, 2019.

\bibitem{smith2017cyclical}
Leslie~N Smith.
\newblock Cyclical learning rates for training neural networks.
\newblock In {\em 2017 IEEE winter conference on applications of computer
  vision (WACV)}, pages 464--472. IEEE, 2017.

\bibitem{sopena1999neural}
Josep~M Sopena, Enrique Romero, and Rene Alquezar.
\newblock Neural networks with periodic and monotonic activation functions: a
  comparative study in classification problems.
\newblock 1999.

\bibitem{sun2017revisiting}
Chen Sun, Abhinav Shrivastava, Saurabh Singh, and Abhinav Gupta.
\newblock Revisiting unreasonable effectiveness of data in deep learning era.
\newblock In {\em Proceedings of the IEEE international conference on computer
  vision}, pages 843--852, 2017.

\bibitem{tolstikhin2021mlp}
Ilya Tolstikhin, Neil Houlsby, Alexander Kolesnikov, Lucas Beyer, Xiaohua Zhai,
  Thomas Unterthiner, Jessica Yung, Andreas Steiner, Daniel Keysers, Jakob
  Uszkoreit, et~al.
\newblock Mlp-mixer: An all-mlp architecture for vision.
\newblock {\em arXiv preprint arXiv:2105.01601}, 2021.

\bibitem{van2021disentangling}
Toon Van~de Maele, Tim Verbelen, Ozan Catal, and Bart Dhoedt.
\newblock Disentangling what and where for 3d object-centric representations
  through active inference.
\newblock {\em arXiv preprint arXiv:2108.11762}, 2021.

\bibitem{vaswani2017attention}
Ashish Vaswani, Noam Shazeer, Niki Parmar, Jakob Uszkoreit, Llion Jones,
  Aidan~N Gomez, {\L}ukasz Kaiser, and Illia Polosukhin.
\newblock Attention is all you need.
\newblock In {\em Advances in neural information processing systems}, pages
  5998--6008, 2017.

\bibitem{wold1987principal}
Svante Wold, Kim Esbensen, and Paul Geladi.
\newblock Principal component analysis.
\newblock {\em Chemometrics and intelligent laboratory systems}, 2(1-3):37--52,
  1987.

\bibitem{wong2002handwritten}
Kwok-wo Wong, Chi-sing Leung, and Sheng-jiang Chang.
\newblock Handwritten digit recognition using multilayer feedforward neural
  networks with periodic and monotonic activation functions.
\newblock In {\em Object recognition supported by user interaction for service
  robots}, volume~3, pages 106--109. IEEE, 2002.

\bibitem{xiao2017/online}
Han Xiao, Kashif Rasul, and Roland Vollgraf.
\newblock Fashion-mnist: a novel image dataset for benchmarking machine
  learning algorithms, 2017.

\bibitem{xu2015show}
Kelvin Xu, Jimmy Ba, Ryan Kiros, Kyunghyun Cho, Aaron Courville, Ruslan
  Salakhudinov, Rich Zemel, and Yoshua Bengio.
\newblock Show, attend and tell: Neural image caption generation with visual
  attention.
\newblock In {\em International conference on machine learning}, pages
  2048--2057. PMLR, 2015.

\bibitem{zhu2007stochastic}
Song-Chun Zhu and David Mumford.
\newblock {\em A stochastic grammar of images}.
\newblock Now Publishers Inc, 2007.

\end{thebibliography}
}

\end{document}